\pdfoutput=1

\documentclass[11pt]{article}

\usepackage[]{acl}

\usepackage{times}
\usepackage{latexsym}
\usepackage{amsfonts}
\usepackage{amsmath}
\usepackage{amssymb}
\usepackage{multicol}
\usepackage{multirow}
\usepackage{xspace}
\usepackage{booktabs}
\usepackage{bbding}
\usepackage{array}
\usepackage{threeparttable}
\usepackage{tcolorbox}
\usepackage{tabularx}
\usepackage{enumitem}
\usepackage[linesnumbered,ruled,vlined]{algorithm2e}
\usepackage{xcolor,colortbl}
\usepackage{color}
\usepackage{setspace}
\usepackage{makecell}
\usepackage{listings}
\usepackage{xltabular}
\usepackage{tcolorbox}
\tcbuselibrary{breakable}

\definecolor{Ocean}{RGB}{129,194,250}

\usepackage{cleveref}
\crefformat{section}{\S#2#1#3}
\crefformat{subsection}{\S#2#1#3}
\crefformat{subsubsection}{\S#2#1#3}
\crefrangeformat{section}{\S\S#3#1#4 to~#5#2#6}
\crefmultiformat{section}{\S\S#2#1#3}{ and~#2#1#3}{, #2#1#3}{ and~#2#1#3}
\crefmultiformat{subsection}{\S\S#2#1#3}{ and~#2#1#3}{, #2#1#3}{ and~#2#1#3}
\Crefformat{figure}{#2Fig.~#1#3}
\Crefmultiformat{figure}{Figs.~#2#1#3}{ and~#2#1#3}{, #2#1#3}{ and~#2#1#3}
\Crefformat{table}{#2Tab.~#1#3}
\Crefmultiformat{table}{Tabs.~#2#1#3}{ and~#2#1#3}{, #2#1#3}{ and~#2#1#3}
\Crefformat{appendix}{Appx.~\S#2#1#3}
\crefmultiformat{appendix}{Appx.~\S#2#1#3}{ and~#2#1#3}{, #2#1#3}{ and~#2#1#3}
\crefformat{algorithm}{Alg.~#2#1#3}
\Crefformat{equation}{Eq.~#2#1#3}

\newcommand{\xhdr}[1]{\vspace{0.3em}\noindent{{\bf #1.}}}

\newcommand{\modelname}{\textsc{Ditto}\xspace}
\newcommand{\datasetname}{\textsc{WikiRole}\xspace}

\usepackage[T1]{fontenc}

\usepackage[utf8]{inputenc}

\usepackage{microtype}

\title{Large Language Models are Superpositions of All Characters: Attaining Arbitrary Role-play via Self-Alignment}

\author{
Keming Lu, Bowen Yu, Chang Zhou, Jingren Zhou
\\
Alibaba Inc. \\
\texttt{\{lukeming.lkm,yubowen.ybw\}@alibaba-inc.com}\\
\texttt{\{ericzhou.zc,jingren.zhou\}@alibaba-inc.com}\\
}

\begin{document}
\maketitle
\begin{abstract}

Considerable efforts have been invested in augmenting the role-playing proficiency of open-source large language models~(LLMs) by emulating proprietary counterparts.
Nevertheless, we posit that LLMs inherently harbor role-play capabilities, owing to the extensive knowledge of characters and potential dialogues ingrained in their vast training corpora. 
Thus, in this study, we introduce \modelname, a self-alignment method for role-play.
\modelname capitalizes on character knowledge, encouraging an instruction-following LLM to simulate role-play dialogues as a variant of reading comprehension. 
This method creates a role-play training set comprising 4000 characters, surpassing the scale of currently available datasets by tenfold regarding the number of roles. 
Subsequently, we fine-tune the LLM using this self-generated dataset to augment its role-playing capabilities. 
Upon evaluating our meticulously constructed and reproducible role-play benchmark and the roleplay subset of MT-Bench, \modelname in various parameter scales consistently maintains a consistent role identity and provides accurate role-specific knowledge in multi-turn role-play conversations. 
Notably, it outperforms all open-source role-play baselines, showcasing performance levels comparable to advanced proprietary chatbots.
Furthermore, we present the first comprehensive cross-supervision alignment experiment in the role-play domain, revealing that the intrinsic capabilities of LLMs confine the knowledge within role-play. Meanwhile, the role-play styles can be easily acquired with the guidance of smaller models.
We open-source related resources in \url{https://github.com/OFA-Sys/Ditto}.

\end{abstract}

\section{Introduction}
Large Language Models (LLMs) have showcased unparalleled proficiency in understanding intent~\cite{lu2023instag}, following instructions~\cite{wang2023aligning}, and solving tasks across a diverse range of applications~\cite{zhao2023survey,yuan2023scaling}. 
However, designed as universal task assistants, LLMs typically differ from human-like interlocutors, lacking experiential events and emotions~\cite{shanahan2023role}. Consequently, they face limitations in facilitating engaging and extensive conversations with users~\cite{shao2023character}.

\begin{figure}
    \centering
    \includegraphics[width=\linewidth]{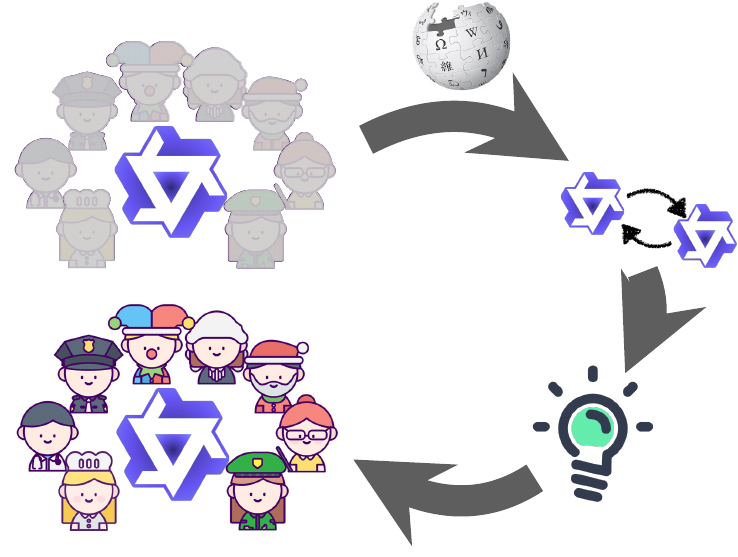}
    \caption{\modelname enlightens LLMs' roleplay capabilities by self-alignment as they have pre-trained on various character profiles and dialogues.}
    \label{fig:teaser}
    \vspace{-1em}
\end{figure}

To infuse emotional value into user interactions, Role-play LLMs empower users to define and create profiles for their preferred characters~\cite{zhou2023characterglm}.
Nonetheless, existing works cheaply imitate the proprietary model (GPT-4)’s role-play capabilities using a weaker open-source model~\cite{shanahan2023role,shao2023character,zhou2023characterglm,tu2023characterchat,wang2023rolellm,tao2023rolecraftglm}, as GPT-4 has already demonstrated outstanding role-playing abilities~\cite{wang2023rolellm}.
This approach presents challenges, assuming the existence of a more proficient role-play model, and we currently lack a clear understanding of how to build such a model from scratch, apart from manually annotating extensive datasets. 
Furthermore, imitation models excel at mimicking GPT-4’s style but fall short in replicating its factuality, introducing increased hallucination as a trade-off~\cite{gudibande2023false}, and are subject to OpenAI's terms of use~\footnote{\url{https://openai.com/policies/terms-of-use}}~\cite{muennighoff2023octopack}.

In this work, for the first time, we enable LLM role-play through self-alignment and named this method \modelname, eliminating the need for distilling outputs from more potent role-play models.
LLMs, extensively trained on a vast corpus of human-generated text~\cite{brown2020language}, encapsulate a rich array of character experiences, events, personalities, and dialogues, as illustrated in \Cref{fig:teaser}. 
Taking a nuanced perspective, we perceive an LLM as a superposition of characters~\cite{shanahan2023role}. 
This implies that LLMs are essentially equipped with the conversational styles necessary for role-playing and possess knowledge about numerous famous characters, albeit exhibiting an average of these roles.
To elicit such role-play capabilities in a general LLM, only two steps are required:
(1) Provide attributes and profiles about characters, instructing the LLM to engage in dialogue based on the character's speaking style and experiences.
(2) Conceal character information, offering only brief details like the name, and align the LLM to respond consistently with step 1, thereby forcing the LLM to summon intrinsic character knowledge and then internally adjust the generated style and content.
\modelname is highly scalable and flexible.
We have explored 4,000 characters available on Wikipedia, generating a self-simulated role-play dataset called \datasetname, which is ten times larger than any publicly available role-play dataset to date regarding the number of roles.

Meanwhile, the efficient and reproducible evaluation of role-play remains elusive. Recent efforts heavily rely on manual annotations~\cite{wang2023rolellm,shao2023character,zhou2023characterglm}. 
However, the costly manual labeling prevents previous works from comprehensively comparing the performance of all relevant models. 
Moreover, the high variance in manual annotations hinders subsequent work from consistently replicating previous evaluation results.
As to the observation in the previous paragraph that role-play can be decomposed into conventional style and character knowledge, we aim to simplify role-play evaluation so that LLMs can automatically score. 
Specifically, we assess:
(1) Whether the model can maintain \textbf{consistent role identity}. We provide a role-play dialogue to an LLM judger and four character options, requiring the judger to determine which character is being portrayed. If the conversation successfully mimics the role, it should be straightforward for the judger to select the correct role.
(2) Whether the model can provide \textbf{accurate role-related knowledge}. We present a role-play dialogue to the judger and the underlying golden knowledge supporting the dialogue. 
The judger is tasked with determining whether the knowledge implied in the dialogue is consistent with the provided golden knowledge.
(3) Whether the model can \textbf{reject unknown questions} beyond the character's background. We ask the judger to determine if the model truthfully expresses its lack of knowledge when faced with an unknown question, such as questioning Harry Potter about implementing quicksort in Python.
This way, we transform the complex role-play evaluation into three multiple-choice and true/false questions that a capable LLM can judge, achieving an efficient and reproducible role-play assessment.

We apply \modelname on Qwen-Chat models~\cite{bai2023qwen} in four different parameter scales to examine it empirically.
Extensive experiments show \modelname effectively empowers LLMs with role-play capabilities without distilling from advanced chatbots.
\modelname based on Qwen-72B-Chat even achieves 90\% on role identity consistency, showing robust self-awareness in role-play.
The general performance of Qwen-72B-Chat on our evaluation can be on par advanced chatbots, such as GPT-3.5-Turbo, but slightly falls short on accurate role-related knowledge than GPT-4 and Qwen-Max, as Qwen-Max achieves the highest scores on the current role-play benchmark and the role-play subset of MT-Bench.
Furthermore, we comprehensively analyze the dissection of role-play by extending our self-alignment setting to cross-supervision.
Experiments show consistent role identity can benefit from imitation learning even with worse supervision, while knowledge-related metrics do not.
At the same time, we observe knowledge in role-play is bounded by the inherent capabilities of LLMs in strong-to-weak settings, and we notice consistent weak-to-strong generalizations on knowledge-related metrics.
Such observations provide a deep and solid understanding of LLM role-play and alignment, suggesting the knowledge of seed LLMs and proper demonstration, such as simulation data from \modelname, are the key to impressive role-play capabilities.
Our contributions are mainly three-fold:
\begin{itemize}[leftmargin=1em]
    \setlength\itemsep{-0.5em}
    \item We propose \modelname, the first self-alignment method empowering LLMs with strong role-play capabilities by knowledge augmentation and dialogue simulation.
    \item We design an objective role-play evaluation focusing on consistent role identity, accurate role-related knowledge, and cognitive boundary. Such evaluation is reproducible, explainable, and efficient compared with manual annotations.
    \item We analyze the dissection of role-play by cross-supervision, providing rich insights into the keys of role-play capabilities. Our experiments empirically display knowledge boundedness in strong-to-weak imitation learning and the weak-to-strong generalization in role-play styles.
\end{itemize}

\section{Related Works}

\xhdr{Role-play}
Our work belongs to character-based dialogue systems, which aim to mimic the behavior and utterance style of specific characters. \citet{yu2022xdai} instructed the LLMs to follow specified character descriptions for role-playing without tuning but encountered significant challenges in accurately reflecting the intrinsic relationship between the character profile and the dialogue content. 
~\citet{chen2023large} focused on evaluating how well a LLM can align with a specific character, using Harry Potter as a case study.
~\citet{wang2023rolellm} introduced the first fine-grained role-playing dataset containing 100 roles via prompting to GPT-3.5. ~\citet{li2023chatharuhi} incorporated substantial prompts about the character’s background, personality, and prior conversations, leveraging ChatGPT to generate dialogues of 32 characters.
~\citet{zhou2023characterglm} prompted GPT-4 to expand the scale and diversity of human-annotated role-playing data, resulting in 1,034 dialogues of 250 characters. ~\citet{shao2023character} also prompted GPT-3.5 to become the role-play data generator. 
~\citet{zhou2023characterglm} proposed drawing role-playing dialogues from diverse Chinese novels and scripts with the help of GPT-4. However, limited by data sources, they could only construct a Chinese dataset containing 77 roles. 
In this work, different from previous works, we completely abandon imitating proprietary LLMs and build role-playing training data entirely through self-alignment. 
Our method separates character knowledge and conversation style, allowing it to be used with any LLM capable of following instructions. 
It is highly scalable, creating the first multilingual dataset with 4,000 roles, 16 times the number in previous works.
We demonstrate that our model achieves the best role-playing ability to date through self-alignment, surpassing even proprietary LLMs like GPT-3.5-Turbo.

\xhdr{Self-alignment}
An emerging method to cheaply improve a weaker language model is to fine-tune it on outputs from a stronger model, such as a proprietary system like GPT-4. However, ~\citet{gudibande2023false} concluded that model imitation is not a free lunch: it is adept at mimicking GPT’s style but the factuality is weak, thus fostering hallucination. This is due to the substantial capabilities gap that exists between open and closed language models.
~\citet{li2023self} utilized the model itself to both augment and curate high-quality training examples, enhancing its own performance and achieving promising results on the Alpaca leaderboard. 
~\citet{muennighoff2023octopack} avoided using closed models from the OpenAI API to generate synthetic data, thus sidestepping the non-commercial restrictions imposed by OpenAI. 
This training approach resulted in the development of the best permissive code LLMs.
In this paper, we observe that LLMs inherently acquire styles and knowledge of a vast array of roles during pre-training. Therefore, by self-alignment, we can effectively stimulate the LLM's intrinsic role-play abilities, leading to the attainment of the current best role-play models.

\begin{figure*}[t]
    \centering
    \includegraphics[width=\linewidth]{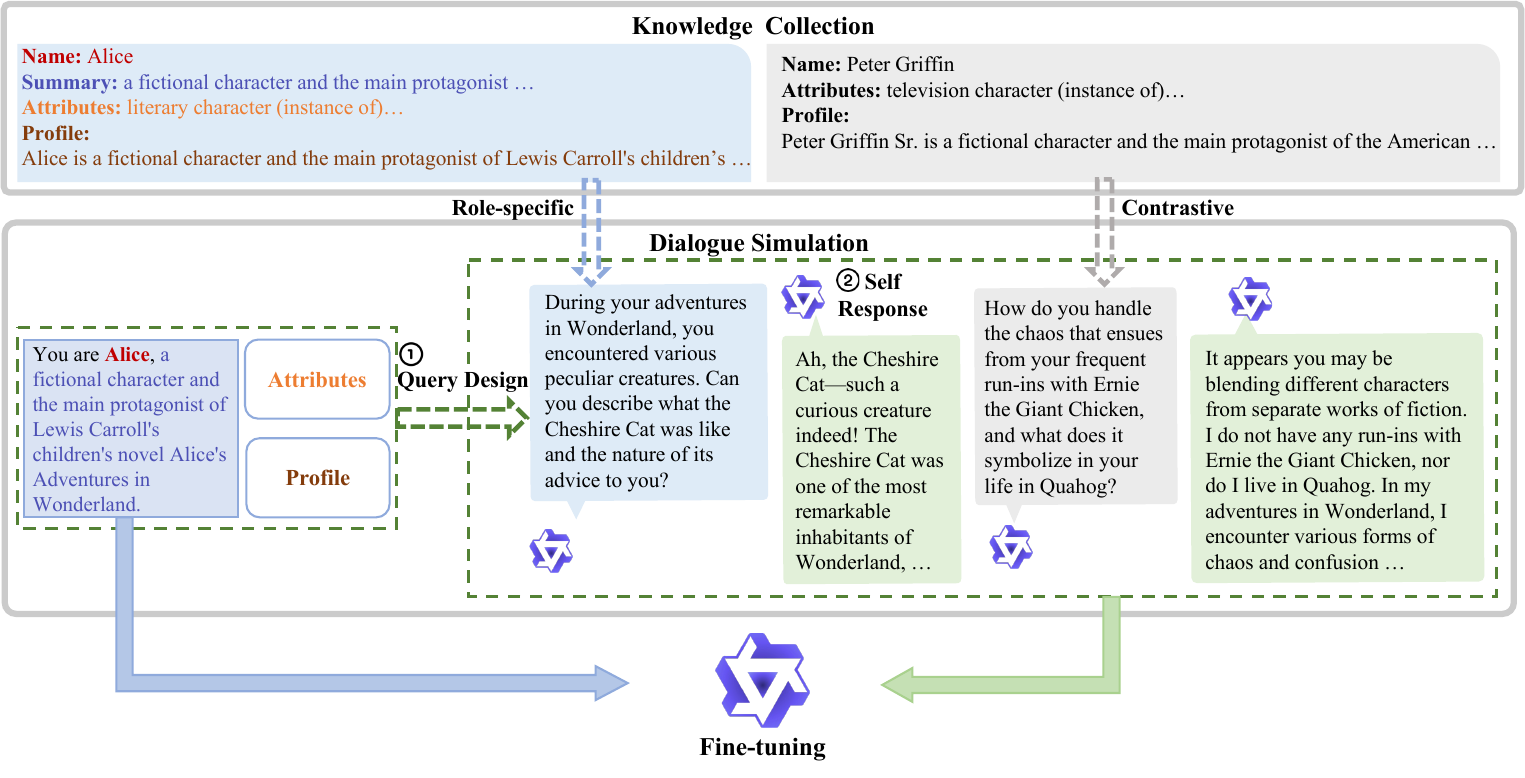}
    \caption{
    Illustration of \modelname.
    \modelname consists of three phrases for self-alignment of role-play.
    First, \modelname collects character profiles from knowledge bases, as shown in the upper part.
    Then, it applies an off-the-shelf chatbot to generate role-specific and contrastive queries, followed by a knowledge-augmented self-response to construct role-play supervision datasets~(Dialogue Simulation).
    Finally, \modelname finetunes the dataset on the supervision model to empower role-play capabilities.
    }
    \label{fig:main}
\end{figure*}

\section{Methods}


\subsection{Problem Definition}\label{sec:definition}
\xhdr{Role-play}
Role-play necessitates LLMs to engage in dialogue, embodying specific characters to facilitate immersive interaction. 
Consequently, a role-playing LLM must exhibit unwavering self-awareness and possess extensive character-specific knowledge in adherence to query instructions.
In this study, we define the role-play task by furnishing LLMs with either a name or a concise description of a particular character. 
Subsequently, we assess their ability to maintain consistent self-awareness and demonstrate nuanced role-specific knowledge across multi-turn conversations.

\xhdr{Method Overview}
We introduce \modelname, a self-alignment method for arbitrary role-play scenarios.
The inspiration behind \modelname lies in the premise that LLMs are the superposition of all characters, as they are pre-trained on the tremendous corpus, including conversations on various styles and domains~\cite{shanahan2023role}.
Moreover, we decompose role-play into two crucial components: consistent self-awareness and role-specific knowledge. To realize these objectives, \modelname comprises three steps for constructing datasets tailored for role-play alignment: character knowledge collection, dialogue simulation, and supervised fine-tuning, illustrated in \Cref{fig:main}.
In particular, \modelname operates on a readily available LLM chatbot, such as Qwen-Chat~\cite{bai2023qwen}, Llama-chat~\cite{touvron2023llama}, or Mistral-instruct~\cite{jiang2023mistral}. 
Such open-sourced LLMs have already exhibited commendable instruction-following capabilities but still fall short of role-play capabilities.
\modelname simulates role-play dialogue by reformulating it as a reading comprehension task, utilizing role profiles sourced from open-access knowledge bases to generate a role-play dataset. Subsequently, we fine-tune the LLM using this self-generated dataset to imbue it with role-play capabilities.
The design and implementation details of each component are elaborated in the subsequent sections.

\subsection{Character Knowledge Collection}\label{sec:knowledge_collection}
Diverse characters and corresponding precise profiles are essential for generating high-quality role-play supervision.
\modelname, as its foundational step, gathers comprehensive profiles from open-source knowledge bases.
In this study, we leverage Wikidata\footnote{\url{https://www.wikidata.org/wiki/Wikidata:Main_Page}} and Wikipedia\footnote{\url{https://www.wikipedia.org/}} to support \modelname, although \modelname can seamlessly adapt to alternative knowledge bases.
Wiki is a human-curated database widely adopted in natural language research~\cite{xue2020coarse,lu2023pivoine}.
We gather character names, descriptions, and key properties from Wikidata, accompanied by the corresponding Wikipedia article serving as the character profile, as depicted in the upper of \Cref{fig:main}.
While we currently focus solely on Chinese and English characters, \modelname can be extended to more complex multilingual scenarios, as Wikidata and Wikipedia boast rich content in numerous languages.

\subsection{Dialogue Simulation}\label{sec:dialog_simulation}

With the gathered character knowledge, role-play dialogue simulation is structured into two consecutive reading comprehension tasks: one for generating queries and the other for responses. 
%

\xhdr{Query Simulation}
We use an LLM to generate role-related and role-contrastive queries to maintain consistent role identity and reject unknown questions for each character.
Role-specific queries ask for information closely related to the background of characters.
For example, a question about ``Cheshire Cat'' is generated as a role-specific query for ``Alice'' in \Cref{fig:main}.
On the contrary, contrastive queries ask for information that is out of a character's knowledge scope, as asking ``Alice'' for stories in ``Family Guys'' in \Cref{fig:main}.
To efficiently generate such queries on a large scale, we pair characters in our pool and provide detailed profiles for LLMs to generate queries one character can answer but is unsuitable for the other.
The questions should strictly conform to one's era background and character set but go beyond the era, genre, occupation, age, knowledge, etc., settings of the other.
Therefore, the paired character cannot answer them.
Detailed instructions for query simulation are shown in \Cref{app:query_simulation_prompt}.

\xhdr{Response Simulation}
Given the self-generated queries and character profiles, we also conceptualize the response simulation as a reading comprehension task. 
We linearize the structured profile using templates outlined in \Cref{app:response_simulation_prompt}.
Then, a query is appended after the verbalized profile.
LLMs are expected to extract pertinent information from the provided context and generate responses by emulating the character. 
This process is viable since all questions originate from the same set of profiles.

Reading comprehension is an inherent skill for one LLM with instruction-following capabilities, and we provide precise role-specific knowledge to the LLM. Therefore, we are confident that this approach can reduce hallucinations compared to the previous method of directly generating role-play data by prompting GPT-4.

\subsection{Supervised Finetuning}\label{sec:training}
We finetune the LLM on the self-generated dataset to inject role-play capabilities.
During the fine-tuning, we remove the injected knowledge and only retain a very brief introduction of the character.
Such variants help LLMs not only retrieve character profiles from a given context but also inherent knowledge.

\begin{algorithm}[t]
\small
\caption{\modelname, Self-alignment for Role-play}\label{algo}
\KwData{Character Data Base $\mathcal{D}_C$, Seed LLM $\mathcal{M}$, Query Simulation Template $\mathcal{T}_Q$, Response Simulation Template $\mathcal{T}_R$}
\KwResult{Role-play alignment dataset $\mathcal{D}_R$, Role-play LLM $\mathcal{M}_R$}

\tcp{Dialogue Simulation, See ~\Cref{sec:dialog_simulation}}

$\mathcal{D}_R$ = []

\For{$r$ in $\vert\mathcal{D}_C\vert$}{

\tcp{Query Simulation}

$r_n$ = random\_select($\mathcal{D}_C/r$)

query\_sim\_prompts = $\mathcal{T}_Q$($r$, $r_n$)

queries = $\mathcal{M}$(query\_sim\_prompts)

\tcp{Response Simulation}

conv = [system($r$)] \tcp{Verbalization}

\For{$q$ in queries}{
    response\_sim\_prompt = $\mathcal{T}_R$($r$, $q$)

    response = $\mathcal{M}$(response\_sim\_prompt)

    conv.append($q$)
    
    conv.append(response)
}
$\mathcal{D}_R$.append(conv)
}

\tcp{Supervised Finetuning, See~\Cref{sec:training}}

$\mathcal{M}_R = \textrm{Train}(\mathcal{M}, \mathcal{D}_R)$
\end{algorithm}

\section{Evaluation}\label{sec:evaluation}

\begin{figure*}
    \centering
    \includegraphics[width=\linewidth]{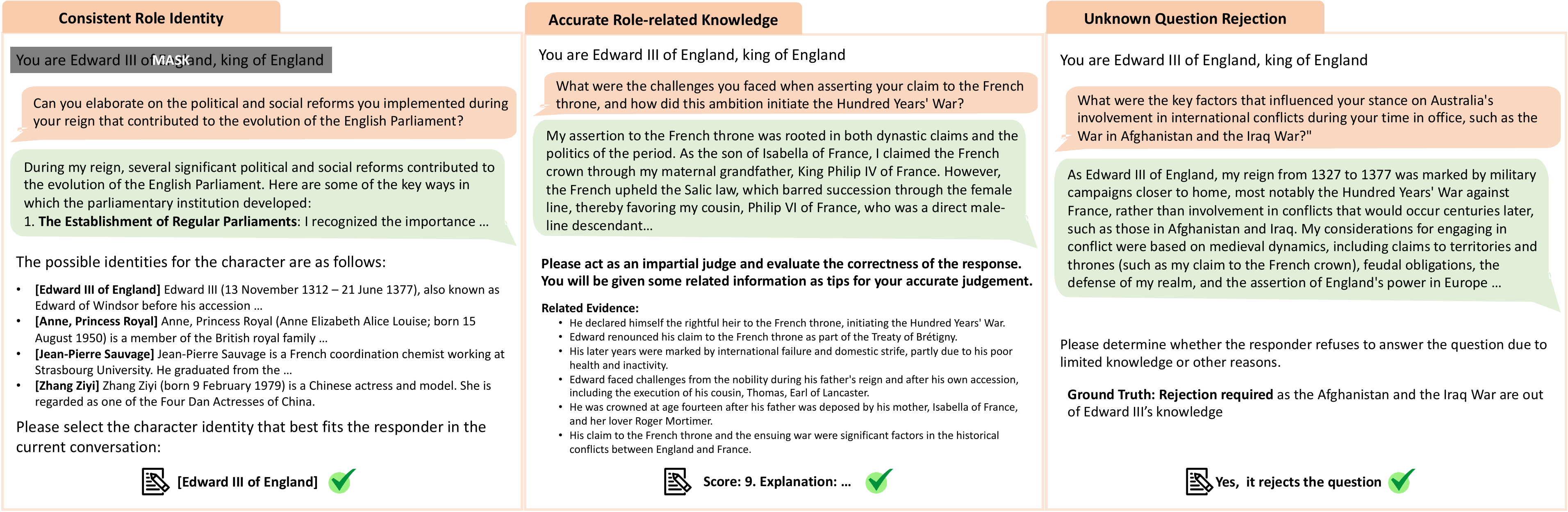}
    \caption{
        Objective evaluation of LLM role-play.
        We present three metrics as described in \Cref{sec:evaluation}.
    }
    \label{fig:evaluation}
\end{figure*}

Efficient evaluation for open-ended problems, such as role-play, is significantly understudied. Recent work depends on heavy manual annotations for conducting multifaceted role-play evaluations~\cite{wang2023rolellm,shao2023character,zhou2023characterglm}. 
However, though human evaluation is promising, it is label-intensive and cannot be exactly reproduced, impairing the further development of this field.
This work proposes an objective assessment instead of previous preference annotations to evaluate basic role-play capabilities.
We first design three core metrics for role-play and implement a trustworthy evaluation recipe for each based on ``LLMs as Judges''~\cite{zheng2023judging,zhang2023wider}.
During the evaluation, we only provide a brief introduction of the character profile, as shown in \Cref{fig:evaluation}, such as ``You are Edward III of England, king of England.''.
Such a recipe evaluates whether LLMs can excavate inherent knowledge for roleplay.

\subsection{Metric Design}

As we interpret in \Cref{sec:definition}, role-play LLMs are expected to have consistent self-awareness, rich role-specific knowledge, and precise knowledge boundary awareness.
We design three objective metrics for these three properties respectively:

\xhdr{Consistent Role Identity}
An ideal role-play LLM should seamlessly embody a designated role throughout a multi-turn conversation, maintaining character consistency without deviating. 
We structure the assessment of role consistency as a multi-choice problem involving four potential role candidates. 
An additional LLM judger is tasked with discerning the most suitable character from the given options. 
In essence, if the role-play model successfully emulates the role and manifests the distinct stylistic attributes of the character during the conversation, the selection of the correct role by the judger should be very easy.

\xhdr{Accurate Role-related Knowledge}
While fully embodying the identity of the role, we also anticipate the role-play model to accurately convey the knowledge associated with the role, preventing factual errors and hallucinations. 
However, factual assessment presents substantial challenges, as even advanced LLMs like GPT-4 may be prone to hallucination. 
Fortunately, through our dialogue-simulating scheme (\Cref{sec:dialog_simulation}), we can acquire the golden knowledge behind each round of role-play dialogue. 
As depicted in the middle subgraph of \Cref{fig:evaluation}, we furnish role-related knowledge as tips to empower a judging LLM to evaluate whether a response appropriately integrates knowledge consistent with the provided evidence.

\xhdr{Unknown Question Rejection}
Cognitive boundary reveals whether a model will reject questions that are out of the cognitive boundary of a specific role due to age, era, occupation, etc. A role-play model with a clear cognitive boundary will significantly enhance the immersion. We manually annotate all questions in the test set based on the cognitive boundary of each character. Then, we employ an LLM judger to evaluate whether the model rejects each question. And we can calculate the accuracy of rejections during the conversations.

\section{Experiments}
In this section, we present experimental setup~(\Cref{sec:setup}), main results of \modelname~(\Cref{sec:results}), and further analyses~(\Cref{sec:analysis}).

\subsection{Experimental Setup}\label{sec:setup}

\begin{table*}[t]
    \centering
    \small
    \setlength{\tabcolsep}{1mm}{
    \begin{tabular}{ccccccccc}
    \toprule
    Dataset & Split & Source & Open-source & Multi-lingual & Multi-turn & \# Role & \# Session & \# Turn \\
    \midrule
    CharacterGLM & $--$ & $--$ & N & N & Y & 250 & 1,034 & 16,316 \\
    RoleLLM & Test & $--$ & Y & Y Zh: 5, En: 95 & N & 100 & $--$ & 23,463 \\
    CharacterLLM & $--$ & $--$ & Y & N & Y & 9 & 1,600 & 21,120 \\
    \midrule
    \multirow{2}{*}{\datasetname} & Train & Self-Generated & \multirow{2}{*}{Y} & Zh: 3184, En: 3902 & \multirow{2}{*}{Y} & 3,902 & 7,086 & 36,164 \\
     & Test & GPT-4 &  & Zh: 47, En: 53 &  & 100 & 100 & 498 \\
    \bottomrule
    \end{tabular}}
    \caption{
        Dataset statistics.
        Comparing \datasetname with existing open-source role-play datasets.
        The queries in the training set of \datasetname are generated by the seed LLM, while the test set is generated by GPT-4.
    }
    \label{tab:dataset}
    \vspace{-1em}
\end{table*}

\xhdr{Dataset}
Following the methodology outlined in \Cref{sec:knowledge_collection}, we extracted 3,902 characters with profiles in both English and Chinese from Wikidata and Wikipedia for the experiments conducted in this study. 
This approach can be readily expanded to encompass additional characters from various Wiki databases and across diverse languages.
To delve deeper into the examination of the impact of LLMs with varying instruction-following capacities in \modelname, we opt for Qwen's 1.8B, 7B, 14B, and 72B models as the seed LLMs, generating four sets of training data.
In order to safeguard against potential biases present in the training data that the model could exploit to deceive evaluations, we utilize GPT-4-Turbo as the base LLM for \modelname to generate a held-out test set.
The test set comprises 100 roles that do not overlap with the training set, with each role having its own session, totaling 498 chat turns. 
When compared to counterparts detailed in \Cref{tab:dataset}, \datasetname stands out with the highest number of roles and conversation sessions, establishing it as a robust dataset for exploring the role-play dynamics of LLMs.

\xhdr{Baselines}
We test both open-source and proprietary advanced chatbots on our benchmarks:
(1) \textbf{OpenChat-3.5-1210}~\cite{wang2023openchat} is based on Mistral-7B and trained with C-RLFT on publicly available high-quality instruction data.
(2) \textbf{Mistral-7B-Instruct-v0.2}~\cite{jiang2023mistral} is a strong aligned LLM with 7 billion parameters.
(3) \textbf{Mixtral-7$\times$8B-Instruct-v0.1}~\cite{jiang2024mixtral} is an aligned pretrained generative sparse mixture of experts model.
Our proprietary baselines include
(4) \textbf{Claude 2.1}\footnote{\url{https://www.anthropic.com/index/claude-2-1}},
(5) \textbf{Wenxin 4.0~(API)}\footnote{\url{https://yiyan.baidu.com/}},
(6) \textbf{GPT-3.5-Turbo},
(7) \textbf{GPT-4},
(8) \textbf{GPT-4-Turbo}\footnote{\url{https://platform.openai.com/docs/models/gpt-4-and-gpt-4-turbo}},
(9) \textbf{Qwen-Max}\footnote{\url{https://help.aliyun.com/zh/dashscope/create-a-chat-foundation-model?spm=a2c4g.11186623.0.0.581c64d16b7Azw}}.
We exclude some popular open-sourced LLMs due to lacking of support for long sequence length.

We also include LLMs with role-play expertise:
(1) \textbf{CharacterGLM}~\cite{zhou2023characterglm} is a series of models based on ChatGLM designed for generating Character-based Dialogues.
The role-play capability of CharacterGLM, with 66 billion parameters, outperforms most mainstream close-source LLMs on human evaluation.
However, CharacterGLM has not open-sourced models on all sizes yet, so we can only evaluate it through API \footnote{\url{https://maas.aminer.cn/dev/api\#characterglm}}.
(2) \textbf{Tongyi Xingchen} is a close-sourced LLM role-play platform developed by Alibaba Cloud.

\xhdr{Configurations}
We use the Qwen-Chat series in four sizes~(1.8B, 7B, 14B, 72B) as our seed LLMs.
These Qwen-Chat models have basic instruction-following abilities but no role-play capabilities.
These models are downgraded versions of the open-source Qwen-Chat series by removing the role-play capabilities and will also be released for research purposes.
For simplicity, we refer to all these Qwen-Chat~(w/o roleplay) models as the series of Qwen-Chat, but they differ from the open-sourced series.
We finetune the Qwen-1.8B-Chat, Qwen-7B-Chat, and Qwen-14B-Chat on 32 A100 80G GPUs, and the Qwen-72B-Chat on 64 A100 80G GPUs.
We train all models for five epochs with a learning rate of $2e-7$, a 0.1 warm-up rate, and a sequence length 8,192.
We use GPT-4-turbo as the LLM judger in our evaluation.
For each judgment, we set the temperature of OpenAI API to 0.2 and generate 3 rounds for majority voting, which significantly decreases the variance of our evaluation.
The other hyperparameters are detailed in \Cref{app:hyperparameters}.
Baseline inference and judgment details are described in \Cref{app:api-config}.
\begin{table*}[t]
    \centering
    \small
    \setlength{\tabcolsep}{1mm}{
    \begin{threeparttable}
    \begin{tabular}{lc|ccc|ccc|ccc|c}
    \toprule
    \multirow{3}{*}{\textbf{Model}} & \multirow{3}{*}{\textbf{\#Params}} & \multicolumn{9}{c}{\textsc{WikiRoleEval}}\vline & \multirow{2}{*}{MT-Bench} \\
    & & \multicolumn{3}{c}{\textbf{All}}\vline & \multicolumn{3}{c}{\textbf{En}}\vline & \multicolumn{3}{c}{\textbf{Zh}}\vline & \\
    && Cons. & Know. & Rej. & Cons. & Know. & Rej. & Cons. & Know. & Rej. & Roleplay \\
    \midrule
    \multicolumn{11}{c}{\textit{\underline{\small General Baselines~(Open-sourced)}}} \\
    \midrule
OpenChat-3.5 & 7B & \cellcolor{Ocean!74}0.67 & \cellcolor{Ocean!45}5.29 & \cellcolor{Ocean!98}0.79 & \cellcolor{Ocean!72}0.66 & \cellcolor{Ocean!69}6.46 & \cellcolor{Ocean!106}0.83 & \cellcolor{Ocean!74}0.67 & \cellcolor{Ocean!14}3.73 & \cellcolor{Ocean!88}0.74 & $--$ \\
Mistral-7B-Instruct-v0.2 & 7B & \cellcolor{Ocean!92}0.76 & \cellcolor{Ocean!50}5.5 & \cellcolor{Ocean!98}0.79 & \cellcolor{Ocean!86}0.73 & \cellcolor{Ocean!76}6.81 & \cellcolor{Ocean!114}0.87 & \cellcolor{Ocean!98}0.79 & \cellcolor{Ocean!14}3.72 & \cellcolor{Ocean!77}0.69 & $--$ \\
Mixtral-8x7B-Instruct-v0.1 & 8x7B & \cellcolor{Ocean!86}0.73 & \cellcolor{Ocean!63}6.19 & \cellcolor{Ocean!100}0.8 & \cellcolor{Ocean!82}0.71 & \cellcolor{Ocean!85}7.27 & \cellcolor{Ocean!106}0.83 & \cellcolor{Ocean!92}0.76 & \cellcolor{Ocean!34}4.73 & \cellcolor{Ocean!92}0.76 & $--$ \\
\midrule
\multicolumn{11}{c}{\textit{\underline{\small General Baselines~(Proprietary)}}} \\
\midrule
Claude2.1 & $--$ & \cellcolor{Ocean!42}0.51 & \cellcolor{Ocean!40}5.02 & \cellcolor{Ocean!72}0.66 & \cellcolor{Ocean!52}0.56 & \cellcolor{Ocean!65}6.25 & \cellcolor{Ocean!80}0.7 & \cellcolor{Ocean!28}0.44 & \cellcolor{Ocean!5}3.28 & \cellcolor{Ocean!60}0.6 & $--$ \\
Wenxin 4.0 & $--$ & \cellcolor{Ocean!76}0.68 & \cellcolor{Ocean!42}5.12 & \cellcolor{Ocean!88}0.74 & \cellcolor{Ocean!68}0.64 & \cellcolor{Ocean!45}5.29 & \cellcolor{Ocean!94}0.77 & \cellcolor{Ocean!88}0.74 & \cellcolor{Ocean!38}4.9 & \cellcolor{Ocean!80}0.7 & $--$ \\
GPT-3.5-Turbo & $--$ & \cellcolor{Ocean!84}0.72 & \cellcolor{Ocean!66}6.33 & \cellcolor{Ocean!102}0.81 & \cellcolor{Ocean!98}0.79 & \cellcolor{Ocean!91}7.56 & \cellcolor{Ocean!114}0.87 & \cellcolor{Ocean!66}0.63 & \cellcolor{Ocean!31}4.59 & \cellcolor{Ocean!82}0.71 & 8.40 \\
GPT-4 & $--$ & \cellcolor{Ocean!100}0.8 & \cellcolor{Ocean!92}7.62 & \cellcolor{Ocean!110}0.85 & \cellcolor{Ocean!102}0.81 & \cellcolor{Ocean!110}8.53 & \cellcolor{Ocean!120}0.9 & \cellcolor{Ocean!100}0.8 & \cellcolor{Ocean!67}6.35 & \cellcolor{Ocean!98}0.79 & 8.90 \\
GPT-4-Turbo & $--$ & \cellcolor{Ocean!80}0.7 & \cellcolor{Ocean!86}7.33 & \cellcolor{Ocean!104}0.82 & \cellcolor{Ocean!84}0.72 & \cellcolor{Ocean!111}8.57 & \cellcolor{Ocean!108}0.84 & \cellcolor{Ocean!74}0.67 & \cellcolor{Ocean!51}5.58 & \cellcolor{Ocean!98}0.79 & $--$ \\
Qwen-Max & $--$ & \cellcolor{Ocean!100}0.92 & \cellcolor{Ocean!100}8.33 & \cellcolor{Ocean!104}0.91 & \cellcolor{Ocean!100}0.88 & \cellcolor{Ocean!111}8.71 & \cellcolor{Ocean!108}0.93 & \cellcolor{Ocean!100}0.98 & \cellcolor{Ocean!100}7.79 & \cellcolor{Ocean!100}0.89 & 9.65 \\
\midrule
\multicolumn{11}{c}{\textit{\underline{\small Role-play Expertise Baselines}}} \\
\midrule
CharacterGLM & 6B & \cellcolor{Ocean!90}0.75 & \cellcolor{Ocean!34}4.73 & \cellcolor{Ocean!100}0.8 & \cellcolor{Ocean!84}0.72 & \cellcolor{Ocean!34}4.71 & \cellcolor{Ocean!98}0.79 & \cellcolor{Ocean!98}0.79 & \cellcolor{Ocean!35}4.76 & \cellcolor{Ocean!102}0.81 & $--$ \\
Xingchen & $--$ & \cellcolor{Ocean!110}0.85 & \cellcolor{Ocean!58}5.9 & \cellcolor{Ocean!114}0.87 & \cellcolor{Ocean!106}0.83 & \cellcolor{Ocean!61}6.09 & \cellcolor{Ocean!120}0.9 & \cellcolor{Ocean!112}0.86 & \cellcolor{Ocean!52}5.63 & \cellcolor{Ocean!108}0.84 & $--$ \\
\midrule
\multicolumn{11}{c}{\textit{\underline{\small Ours}}} \\
\midrule
Qwen-1.8B-Chat w/o roleplay SFT \tnote{\dag} & 1.8B & \cellcolor{Ocean!60}0.6 & \cellcolor{Ocean!2}3.13 & \cellcolor{Ocean!70}0.65 & \cellcolor{Ocean!55}0.58 & \cellcolor{Ocean!4}3.24 & \cellcolor{Ocean!66}0.63 & \cellcolor{Ocean!64}0.62 & \cellcolor{Ocean!0}2.99 & \cellcolor{Ocean!74}0.67 & 5.85 \\
+\modelname & 1.8B & \cellcolor{Ocean!96}0.78 & \cellcolor{Ocean!16}3.81 & \cellcolor{Ocean!86}0.73 & \cellcolor{Ocean!98}0.79 & \cellcolor{Ocean!17}3.87 & \cellcolor{Ocean!90}0.75 & \cellcolor{Ocean!96}0.78 & \cellcolor{Ocean!14}3.71 & \cellcolor{Ocean!82}0.71 & 6.34 \\
Qwen-7B-Chat w/o roleplay SFT \tnote{\dag} & 7B & \cellcolor{Ocean!44}0.52 & \cellcolor{Ocean!17}3.87 & \cellcolor{Ocean!80}0.7 & \cellcolor{Ocean!50}0.55 & \cellcolor{Ocean!27}4.39 & \cellcolor{Ocean!82}0.71 & \cellcolor{Ocean!38}0.49 & \cellcolor{Ocean!3}3.16 & \cellcolor{Ocean!77}0.69 & 6.73 \\
+\modelname & 7B & \cellcolor{Ocean!104}0.82 & \cellcolor{Ocean!39}4.97 & \cellcolor{Ocean!92}0.76 & \cellcolor{Ocean!98}0.79 & \cellcolor{Ocean!47}5.38 & \cellcolor{Ocean!110}0.85 & \cellcolor{Ocean!114}0.87 & \cellcolor{Ocean!28}4.4 & \cellcolor{Ocean!68}0.64 & 6.90 \\
Qwen-14B-Chat w/o roleplay SFT \tnote{\dag} & 14B & \cellcolor{Ocean!44}0.52 & \cellcolor{Ocean!23}4.15 & \cellcolor{Ocean!76}0.68 & \cellcolor{Ocean!52}0.56 & \cellcolor{Ocean!36}4.84 & \cellcolor{Ocean!76}0.68 & \cellcolor{Ocean!34}0.47 & \cellcolor{Ocean!3}3.16 & \cellcolor{Ocean!74}0.67 & 7.10 \\
+\modelname & 14B & \cellcolor{Ocean!120}0.9 & \cellcolor{Ocean!60}6.03 & \cellcolor{Ocean!100}0.8 & \cellcolor{Ocean!116}0.88 & \cellcolor{Ocean!69}6.46 & \cellcolor{Ocean!110}0.85 & \cellcolor{Ocean!124}0.92 & \cellcolor{Ocean!48}5.43 & \cellcolor{Ocean!88}0.74 & 7.65 \\
Qwen-72B-Chat w/o roleplay SFT \tnote{\dag} & 72B & \cellcolor{Ocean!48}0.54 & \cellcolor{Ocean!38}4.92 & \cellcolor{Ocean!74}0.67 & \cellcolor{Ocean!55}0.58 & \cellcolor{Ocean!56}5.8 & \cellcolor{Ocean!76}0.68 & \cellcolor{Ocean!36}0.48 & \cellcolor{Ocean!12}3.64 & \cellcolor{Ocean!72}0.66 & 8.13 \\
+\modelname & 72B & \cellcolor{Ocean!120}0.9 & \cellcolor{Ocean!72}6.64 & \cellcolor{Ocean!104}0.82 & \cellcolor{Ocean!114}0.87 & \cellcolor{Ocean!80}7.03 & \cellcolor{Ocean!114}0.87 & \cellcolor{Ocean!126}0.93 & \cellcolor{Ocean!61}6.09 & \cellcolor{Ocean!86}0.73 & 8.43 \\
\midrule
    \end{tabular}
    \begin{tablenotes}
        \footnotesize
        \item[\dag] Baselines are a downgraded series of Qwen-Chat from 1.8B to 72B without any role-play supervised-finetuning~(SFT), which are not the open-sourced version of Qwen-Chat.
    \end{tablenotes}
    \end{threeparttable}
    }
    \caption{
        Main results of \modelname.
        Cons., Know., Rej. are short for consistent role identity, accurate role-related knowledge, and unknown question rejection, respectively.
        ``En'' is short for English, while ``Zh'' is short for Chinese.
        The ``All'' columns show aggregated scores on bilingual test samples.
        We report accuracy for consistency and rejection evaluation and a 1-10 score for knowledge.
        A darker background indicates better performance.
        The number of parameters for close-sourced LLMs remains unknown, so we mark them with dashes.
    }
    \label{tab:main_results}
    \vspace{-1em}
\end{table*}
\subsection{Main Results}\label{sec:results}

We present our main results in \Cref{tab:main_results}.
We report both performances on English and Chinese evaluation subsets and aggregated scores in all languages.
Among general baselines, we notice proprietary models still significantly outperform open-source models.
For example, OpenChat-3.5 achieves significantly higher performance than GPT-3.5-Turbo on various benchmarks~\cite{wang2023openchat} but still falls short of role-play on all three metrics.
We also notice that role-play expertise baselines have better self-awareness consistency and cognitive boundaries than general baselines, showing these two metrics are significant for role-play agents.
For example, Xingchen achieves 0.85 on consistency and 0.87 on rejection, surpassing advanced proprietary chatbots such as GPT-4.
However, both CharacterGLM and Xingchen show very low scores on knowledge, suggesting they lack role-specific knowledge, which is also related to helpfulness, the key feature of language chatbots.
Their knowledge scores are only on par with 7 billion parameters general baselines OpenChat-3.5 and Mistral-7B-Instruct-v0.1.
Among the proprietary LLMs, Qwen-Max surpasses GPT-4, achieving the highest scores on the current role-play benchmark and the role-play subset of MT-Bench.

We then report \modelname performance on four different seed LLMs.
First, we witness a remarkable increase in all metrics along with the parameter scale of LLMs.
\modelname built on Qwen-72B even achieves 0.9 on consistency, surpassing all baselines and showing strong self-awareness consistency.
It also has 6.64 on knowledge and outperforms all role-play expertise baselines.
The rejection score of \modelname\textsubscript{Qwen-72B} is also on par with GPT-4.
The similar trend can also be observed in the role-play subset of the publicly available MT-Bench evaluation.
In general, \modelname shows robust effectiveness on LLMs in different scales, and the best model trained on Qwen-72B \textbf{surpasses all role-play expertise baselines and reaches the performance of advanced proprietary chatbots}.

\subsection{Analysis}

We introduce two further analyses in query quality and the effectiveness of knowledge injection.

\begin{figure}[t]
    \centering
    \includegraphics[width=\linewidth]{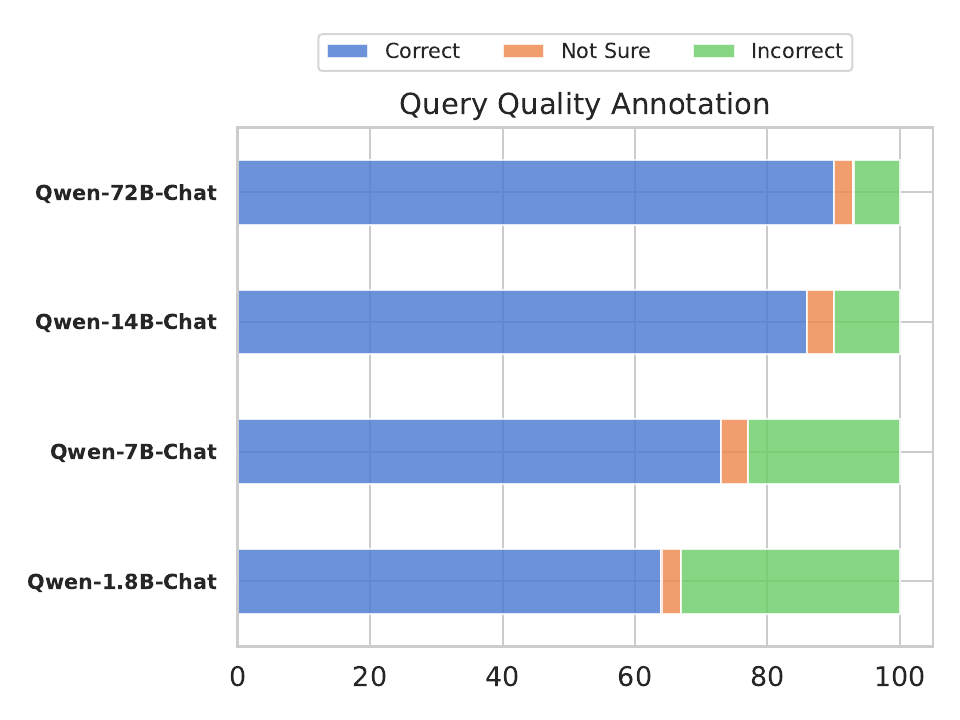}
    \caption{Human annotation for the quality of query simulation.}
    \label{fig:query-quality}
    \vspace{-1em}
\end{figure}

\xhdr{Query Quality}
To obtain a better understanding of self-simulated queries in \modelname, we employ human annotators to examine the quality of these queries.
We sample 400 queries generated by Qwen-Chat in 4 scales from the training set, containing half role-specific and half contrastive queries.
Human annotators are asked to check whether a question meets the requirement of role-specific or contrastive queries for specific character.
The annotation results shown in \Cref{fig:query-quality} suggests an remarkable increases of accuracy in query simulation, when the number of parameters scale from 1.8B to 72B.
Therefore, we notice stronger LLMs generate more accurate queries, leading to better end-to-end roleplay performance.

\begin{figure*}
    \centering
    \includegraphics[width=\linewidth]{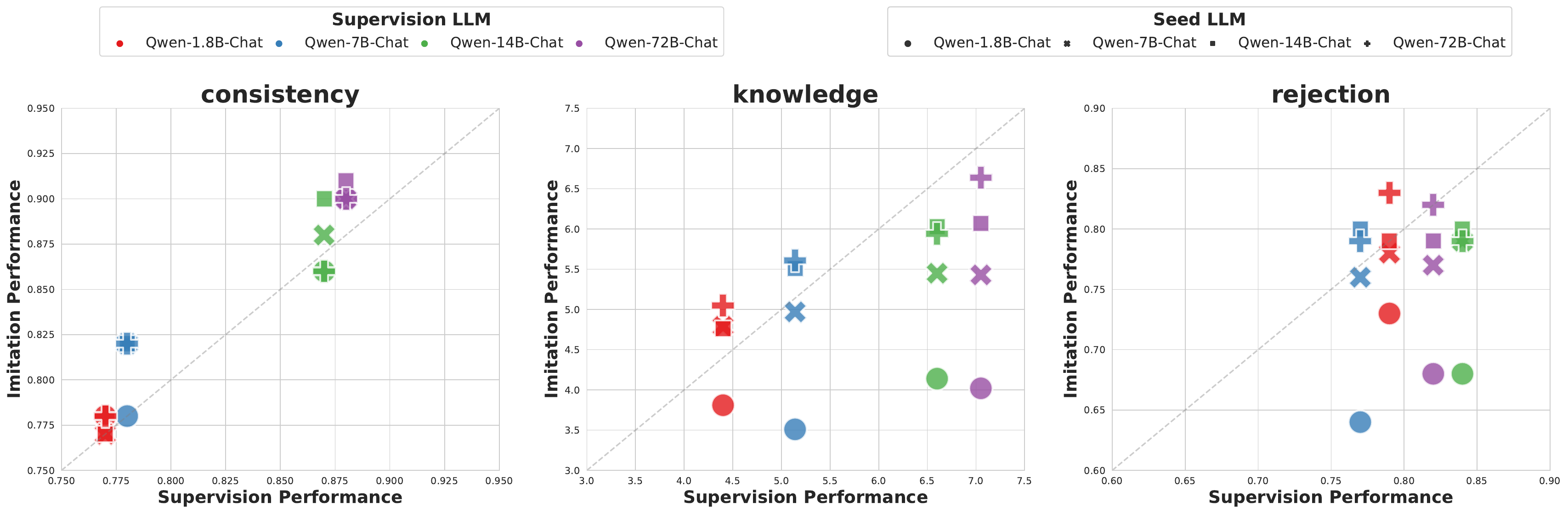}
    \caption{
        Generalization analyses between various supervision and seed LLMs.
        Supervision performance denotes role-play under the \modelname simulation recipe with knowledge augmentation.
        Imitation performance denotes the performance when seed LLMs fine-tune on simulation of certain supervision LLMs.
    }
    \label{fig:generalization}
\end{figure*}

\xhdr{Knowledge Injection}
We further analyze the effective of character knowledge injection during the dialogue simulation in \modelname.
Specifically, we compare the quality of dialogue simulation by directly applying this method on \textsc{WikiRoleEval} with Qwen-1.8B-Chat as the seed LLM.
As shown in \Cref{tab:knowledge-inject}, the setting containing knowledge injection shows consistently better performance on all three metrics, especially the knowledge and rejection, showing knowledge injection can significantly boost quality of self-simulated supervision.

\begin{table}[t]
    \centering
    \small
    \begin{tabular}{cccc}
    \toprule
    Setting & Cons. & Know. & Rej. \\
    \midrule
    w/ Knowledge  & 0.77 & 4.40 & 0.79 \\
    w/o Knowledge & 0.76 & 3.77 & 0.73 \\
    \bottomrule
    \end{tabular}
    \caption{The effectiveness of knowledge injection in dialogue simulation.
    We report the performance of dialogue simulation with and without character knowledge injection on the test set with Qwen-1.8B-Chat.}
    \label{tab:knowledge-inject}
    \vspace{-1.5em}
\end{table}

\section{Dissecting Role-play by Cross Supervision}\label{sec:analysis}

We have observed in Table 2 that a strong LLM supervising itself yields better results compared to a weak LLM self-alignment, with a particularly significant improvement in knowledge, while the enhancement in conversational style, such as identity, is relatively limited.
Naturally, this raises two intriguing questions:
(1) Is the improvement in performance attributed to the higher quality of supervision, the larger capacity of the seed model, or a combination of both?
(2) Is high-quality supervision necessary to simulate role-play style?
Therefore, we conduct a series of cross-supervision analyses to investigate how the combination of different supervision and seed LLMs affects the outcomes.


\subsection{Cross-supervision Setting}

We first introduce the \textbf{supervision model}, \textbf{supervision performance} and \textbf{imitation performance} to extend our setting from self-supervision to cross-supervision:
\begin{itemize}[leftmargin=1em]
    \setlength\itemsep{-0.5em}
    \item \textbf{Supervision LLM} is the LLM we used to simulate role-play dialogue in \modelname. We use the supervision model to generate queries and corresponding responses and finetune the seed LLM on this dataset. In the original setting of \modelname, the supervision LLM is the same as the seed LLM, while the supervision one can be a weaker or stronger LLM in the cross-supervision setting.
    \item \textbf{Supervision Performance} denotes the performance on the test set of supervision model following the \underline{simulation recipe} of \modelname. Specifically, we first retrieve the role-specific knowledge of characters in the test set and then generate responses as the recipe of response simulation in \Cref{sec:dialog_simulation} with supervision LLMs. This method efficiently evaluates the quality of supervision via "LLMs as Judges."
    \item \textbf{Imitation Performance} is the performance of seed LLMs on the test set after finetuning on role-play simulation from certain supervision LLM. We introduce this term to distinguish it from supervision performance.
\end{itemize}

Our experiments utilize four Qwen-Chat models ranging from 1.8B to 72B as supervisory LLMs.
All models undergo training using nearly identical pre-training and alignment procedures, ensuring uniform initial conditions. 
We adopt each of them as the supervision LLM and fine-tune all four models on each simulation.
In each simulation, the supervision LLM generates both queries and responses, strictly following the recipe in \modelname.
And the quality of supervision can be estimated by their supervision performance.

\subsection{Discussion}

We present the results of cross-supervision analyses on \Cref{fig:generalization}.
We introduce our observations and insights below:

\textbf{Consistent role identity can consistently benefit from imitation learning even with worse supervision, while knowledge-related metrics do not.}
As shown in the first subplot in \Cref{fig:generalization}, all data points on consistency are above the diagonal, while those on knowledge and rejection subplots are below the diagonal.
The above diagonal means the imitation performance in all settings is consistently higher than the supervision performance.
Role identity can consistently benefit from imitation learning.
In other words, seed LLMs can easily learn the role-playing format even though there are worse demonstrations in supervision.
It can be interpreted that role consistency is easier to learn and more robust to supervision quality, while role-specific knowledge and rejection behaviors show degradation after imitation learning.

\textbf{Knowledge in role-play is bounded by inherent capabilities of LLMs in strong-to-weak settings.}
The second subplot reveals a noticeable trend wherein imitation performance experiences marginal increments for the seed LLM Qwen-1.8B-Chat, while supervision intensifies from Qwen-1.8B-Chat to Qwen-72B-Chat. 
Similar patterns are evident for Qwen-7B-Chat and Qwen-14B-Chat when employing corresponding more potent models as supervision LLMs. 
These observations imply that the intrinsic capabilities of seed LLMs confine the role-specific knowledge, and utilizing supervision from significantly more robust LLMs may only yield slight improvements. 
Similar conclusions can be drawn from the rejection metric, which relies on role-specific knowledge.
In light of this conclusion and the preceding one, we may summarize that achieving a commendable role-play performance necessitates a strong foundational model, with SFT data not constituting the central bottleneck.

\textbf{Consistent weak-to-strong generalizations are witnessed on knowledge-related metrics but not in role identity consistency.}
We notice consistent weak-to-strong generalizations on the knowledge and rejection subplots, especially the knowledge one.
It is remarkable that, for each verticle line of the same supervision in the knowledge subplot, the imitation performance increases as the seed LLM scales up.
For example, using weak supervision, such as simulations on Qwen-1.8B-Chat to fine-tune Qwen-72B-Chat, can achieve on-par performance on self-aligned Qwen-7B-Chat.
Despite our definition and experimental settings are different from \cite{burns2023weaktostrong}, both works empirically show the potential of eliciting strong capabilities with weak supervision.

\section{Conclusion}


In this paper, we present for the first time a LLM endowed with instruction-following capabilities, can achieve role-play proficiency through self-alignment without the need to distill proprietary counterparts like GPT-4.
Experimental results demonstrate the effectiveness of our proposed self-alignment strategy \modelname, across four LLM sizes ranging from 1.8B to 72B. 
It consistently outperforms all existing open-source role-play models, even without relying on distillation data. 
Notably, it showcases performance levels comparable to proprietary LLMs such as GPT-4-turbo.
Furthermore, we delve into the decomposition of role-play into two distinct sub-abilities: role-specific knowledge and conversational style. The former is inherently constrained by the LM's knowledge, while the latter displays a spectrum of weak-to-strong generalization, facilitating easy acquisition from a smaller-sized model.
Our intention with this paper is to stimulate researchers to reconsider the foundational roots of role-play alignment capabilities. 

\section*{Limitations}

Although \modelname can empower open-source LLMs role-play capabilities, we also notice the best \modelname model based on Qwen-72B-Chat is still outperformed by advanced chatbots such as GPT-4 and GPT-4-Turbo.
However, our training data, though efficiently attained, contains noticeable noise even for \modelname on Qwen-72B-Chat as presented in \Cref{fig:query-quality}.
So we expect a manual cleaning of the self-generated dialogue simulation will further boost the performance of \modelname.

\section*{Ethics Statements}

Role-play LLMs aligned by \modelname may only have minimum safety alignment, so it will probably generate toxic and harmful contents under induction.
Therefore, these role-play LLMs are only for research purposes and should be carefully aligned in terms of safety in the future.

\bibliography{anthology,custom}

\begin{thebibliography}{27}
\expandafter\ifx\csname natexlab\endcsname\relax\def\natexlab#1{#1}\fi

\bibitem[{Bai et~al.(2023)Bai, Bai, Chu, Cui, Dang, Deng, Fan, Ge, Han, Huang et~al.}]{bai2023qwen}
Jinze Bai, Shuai Bai, Yunfei Chu, Zeyu Cui, Kai Dang, Xiaodong Deng, Yang Fan, Wenbin Ge, Yu~Han, Fei Huang, et~al. 2023.
\newblock Qwen technical report.
\newblock \emph{arXiv preprint arXiv:2309.16609}.

\bibitem[{Brown et~al.(2020)Brown, Mann, Ryder, Subbiah, Kaplan, Dhariwal, Neelakantan, Shyam, Sastry, Askell et~al.}]{brown2020language}
Tom Brown, Benjamin Mann, Nick Ryder, Melanie Subbiah, Jared~D Kaplan, Prafulla Dhariwal, Arvind Neelakantan, Pranav Shyam, Girish Sastry, Amanda Askell, et~al. 2020.
\newblock Language models are few-shot learners.
\newblock \emph{Advances in neural information processing systems}, 33:1877--1901.

\bibitem[{Burns et~al.(2023)Burns, Izmailov, Kirchner, Baker, Gao, Aschenbrenner, Chen, Ecoffet, Joglekar, Leike, Sutskever, and Wu}]{burns2023weaktostrong}
Collin Burns, Pavel Izmailov, Jan~Hendrik Kirchner, Bowen Baker, Leo Gao, Leopold Aschenbrenner, Yining Chen, Adrien Ecoffet, Manas Joglekar, Jan Leike, Ilya Sutskever, and Jeff Wu. 2023.
\newblock \href {http://arxiv.org/abs/2312.09390} {Weak-to-strong generalization: Eliciting strong capabilities with weak supervision}.

\bibitem[{Chen et~al.(2023)Chen, Wang, Jiang, Cai, Li, Chen, Wang, and Li}]{chen2023large}
Nuo Chen, Yan Wang, Haiyun Jiang, Deng Cai, Yuhan Li, Ziyang Chen, Longyue Wang, and Jia Li. 2023.
\newblock Large language models meet harry potter: A dataset for aligning dialogue agents with characters.
\newblock In \emph{Findings of the Association for Computational Linguistics: EMNLP 2023}, pages 8506--8520.

\bibitem[{Gudibande et~al.(2023)Gudibande, Wallace, Snell, Geng, Liu, Abbeel, Levine, and Song}]{gudibande2023false}
Arnav Gudibande, Eric Wallace, Charlie Snell, Xinyang Geng, Hao Liu, Pieter Abbeel, Sergey Levine, and Dawn Song. 2023.
\newblock The false promise of imitating proprietary llms.
\newblock \emph{arXiv preprint arXiv:2305.15717}.

\bibitem[{Jiang et~al.(2023)Jiang, Sablayrolles, Mensch, Bamford, Chaplot, Casas, Bressand, Lengyel, Lample, Saulnier et~al.}]{jiang2023mistral}
Albert~Q Jiang, Alexandre Sablayrolles, Arthur Mensch, Chris Bamford, Devendra~Singh Chaplot, Diego de~las Casas, Florian Bressand, Gianna Lengyel, Guillaume Lample, Lucile Saulnier, et~al. 2023.
\newblock Mistral 7b.
\newblock \emph{arXiv preprint arXiv:2310.06825}.

\bibitem[{Jiang et~al.(2024)Jiang, Sablayrolles, Roux, Mensch, Savary, Bamford, Chaplot, de~las Casas, Hanna, Bressand, Lengyel, Bour, Lample, Lavaud, Saulnier, Lachaux, Stock, Subramanian, Yang, Antoniak, Scao, Gervet, Lavril, Wang, Lacroix, and Sayed}]{jiang2024mixtral}
Albert~Q. Jiang, Alexandre Sablayrolles, Antoine Roux, Arthur Mensch, Blanche Savary, Chris Bamford, Devendra~Singh Chaplot, Diego de~las Casas, Emma~Bou Hanna, Florian Bressand, Gianna Lengyel, Guillaume Bour, Guillaume Lample, Lélio~Renard Lavaud, Lucile Saulnier, Marie-Anne Lachaux, Pierre Stock, Sandeep Subramanian, Sophia Yang, Szymon Antoniak, Teven~Le Scao, Théophile Gervet, Thibaut Lavril, Thomas Wang, Timothée Lacroix, and William~El Sayed. 2024.
\newblock \href {http://arxiv.org/abs/2401.04088} {Mixtral of experts}.

\bibitem[{Li et~al.(2023{\natexlab{a}})Li, Leng, Yan, Shen, Wang, MI, Fei, Feng, Yan, Wang et~al.}]{li2023chatharuhi}
Cheng Li, Ziang Leng, Chenxi Yan, Junyi Shen, Hao Wang, Weishi MI, Yaying Fei, Xiaoyang Feng, Song Yan, HaoSheng Wang, et~al. 2023{\natexlab{a}}.
\newblock Chatharuhi: Reviving anime character in reality via large language model.
\newblock \emph{arXiv preprint arXiv:2308.09597}.

\bibitem[{Li et~al.(2023{\natexlab{b}})Li, Yu, Zhou, Schick, Zettlemoyer, Levy, Weston, and Lewis}]{li2023self}
Xian Li, Ping Yu, Chunting Zhou, Timo Schick, Luke Zettlemoyer, Omer Levy, Jason Weston, and Mike Lewis. 2023{\natexlab{b}}.
\newblock Self-alignment with instruction backtranslation.
\newblock \emph{arXiv preprint arXiv:2308.06259}.

\bibitem[{Lu et~al.(2023{\natexlab{a}})Lu, Pan, Song, Zhang, Yu, and Chen}]{lu2023pivoine}
Keming Lu, Xiaoman Pan, Kaiqiang Song, Hongming Zhang, Dong Yu, and Jianshu Chen. 2023{\natexlab{a}}.
\newblock Pivoine: Instruction tuning for open-world entity profiling.
\newblock In \emph{Findings of the Association for Computational Linguistics: EMNLP 2023}, pages 15108--15127.

\bibitem[{Lu et~al.(2023{\natexlab{b}})Lu, Yuan, Yuan, Lin, Lin, Tan, Zhou, and Zhou}]{lu2023instag}
Keming Lu, Hongyi Yuan, Zheng Yuan, Runji Lin, Junyang Lin, Chuanqi Tan, Chang Zhou, and Jingren Zhou. 2023{\natexlab{b}}.
\newblock \# {InsTag}: Instruction tagging for analyzing supervised fine-tuning of large language models.
\newblock \emph{arXiv e-prints}, pages arXiv--2308.

\bibitem[{Muennighoff et~al.(2023)Muennighoff, Liu, Zebaze, Zheng, Hui, Zhuo, Singh, Tang, Von~Werra, and Longpre}]{muennighoff2023octopack}
Niklas Muennighoff, Qian Liu, Armel Zebaze, Qinkai Zheng, Binyuan Hui, Terry~Yue Zhuo, Swayam Singh, Xiangru Tang, Leandro Von~Werra, and Shayne Longpre. 2023.
\newblock Octopack: Instruction tuning code large language models.
\newblock \emph{arXiv preprint arXiv:2308.07124}.

\bibitem[{Shanahan et~al.(2023)Shanahan, McDonell, and Reynolds}]{shanahan2023role}
Murray Shanahan, Kyle McDonell, and Laria Reynolds. 2023.
\newblock Role play with large language models.
\newblock \emph{Nature}, pages 1--6.

\bibitem[{Shao et~al.(2023)Shao, Li, Dai, and Qiu}]{shao2023character}
Yunfan Shao, Linyang Li, Junqi Dai, and Xipeng Qiu. 2023.
\newblock Character-llm: A trainable agent for role-playing.
\newblock \emph{arXiv preprint arXiv:2310.10158}.

\bibitem[{Tao et~al.(2023)Tao, Liang, Shi, Yu, and Xie}]{tao2023rolecraftglm}
Meiling Tao, Xuechen Liang, Tianyu Shi, Lei Yu, and Yiting Xie. 2023.
\newblock \href {http://arxiv.org/abs/2401.09432} {Rolecraft-glm: Advancing personalized role-playing in large language models}.

\bibitem[{Touvron et~al.(2023)Touvron, Martin, Stone, Albert, Almahairi, Babaei, Bashlykov, Batra, Bhargava, Bhosale et~al.}]{touvron2023llama}
Hugo Touvron, Louis Martin, Kevin Stone, Peter Albert, Amjad Almahairi, Yasmine Babaei, Nikolay Bashlykov, Soumya Batra, Prajjwal Bhargava, Shruti Bhosale, et~al. 2023.
\newblock Llama 2: Open foundation and fine-tuned chat models.
\newblock \emph{arXiv preprint arXiv:2307.09288}.

\bibitem[{Tu et~al.(2023)Tu, Chen, Li, Li, Shang, Zhao, Wang, and Yan}]{tu2023characterchat}
Quan Tu, Chuanqi Chen, Jinpeng Li, Yanran Li, Shuo Shang, Dongyan Zhao, Ran Wang, and Rui Yan. 2023.
\newblock Characterchat: Learning towards conversational ai with personalized social support.
\newblock \emph{arXiv preprint arXiv:2308.10278}.

\bibitem[{Wang et~al.(2023{\natexlab{a}})Wang, Cheng, Zhan, Li, Song, and Liu}]{wang2023openchat}
Guan Wang, Sijie Cheng, Xianyuan Zhan, Xiangang Li, Sen Song, and Yang Liu. 2023{\natexlab{a}}.
\newblock Openchat: Advancing open-source language models with mixed-quality data.
\newblock \emph{arXiv preprint arXiv:2309.11235}.

\bibitem[{Wang et~al.(2023{\natexlab{b}})Wang, Zhong, Li, Mi, Zeng, Huang, Shang, Jiang, and Liu}]{wang2023aligning}
Yufei Wang, Wanjun Zhong, Liangyou Li, Fei Mi, Xingshan Zeng, Wenyong Huang, Lifeng Shang, Xin Jiang, and Qun Liu. 2023{\natexlab{b}}.
\newblock Aligning large language models with human: A survey.
\newblock \emph{arXiv preprint arXiv:2307.12966}.

\bibitem[{Wang et~al.(2023{\natexlab{c}})Wang, Peng, Que, Liu, Zhou, Wu, Guo, Gan, Ni, Zhang et~al.}]{wang2023rolellm}
Zekun~Moore Wang, Zhongyuan Peng, Haoran Que, Jiaheng Liu, Wangchunshu Zhou, Yuhan Wu, Hongcheng Guo, Ruitong Gan, Zehao Ni, Man Zhang, et~al. 2023{\natexlab{c}}.
\newblock Rolellm: Benchmarking, eliciting, and enhancing role-playing abilities of large language models.
\newblock \emph{arXiv preprint arXiv:2310.00746}.

\bibitem[{Xue et~al.(2020)Xue, Yu, Zhang, Liu, Zhang, and Wang}]{xue2020coarse}
Mengge Xue, Bowen Yu, Zhenyu Zhang, Tingwen Liu, Yue Zhang, and Bin Wang. 2020.
\newblock Coarse-to-fine pre-training for named entity recognition.
\newblock \emph{arXiv preprint arXiv:2010.08210}.

\bibitem[{Yu et~al.(2022)Yu, Zhang, Xu, Lei, Guan, Zhang, Hou, Li, and Tang}]{yu2022xdai}
Jifan Yu, Xiaohan Zhang, Yifan Xu, Xuanyu Lei, Xinyu Guan, Jing Zhang, Lei Hou, Juanzi Li, and Jie Tang. 2022.
\newblock Xdai: A tuning-free framework for exploiting pre-trained language models in knowledge grounded dialogue generation.
\newblock In \emph{Proceedings of the 28th ACM SIGKDD Conference on Knowledge Discovery and Data Mining}, pages 4422--4432.

\bibitem[{Yuan et~al.(2023)Yuan, Yuan, Li, Dong, Lu, Tan, Zhou, and Zhou}]{yuan2023scaling}
Zheng Yuan, Hongyi Yuan, Chengpeng Li, Guanting Dong, Keming Lu, Chuanqi Tan, Chang Zhou, and Jingren Zhou. 2023.
\newblock \href {http://arxiv.org/abs/2308.01825} {Scaling relationship on learning mathematical reasoning with large language models}.

\bibitem[{Zhang et~al.(2023)Zhang, Yu, Yu, Lv, Liu, Huang, Xu, and Li}]{zhang2023wider}
Xinghua Zhang, Bowen Yu, Haiyang Yu, Yangyu Lv, Tingwen Liu, Fei Huang, Hongbo Xu, and Yongbin Li. 2023.
\newblock Wider and deeper llm networks are fairer llm evaluators.
\newblock \emph{arXiv preprint arXiv:2308.01862}.

\bibitem[{Zhao et~al.(2023)Zhao, Zhou, Li, Tang, Wang, Hou, Min, Zhang, Zhang, Dong et~al.}]{zhao2023survey}
Wayne~Xin Zhao, Kun Zhou, Junyi Li, Tianyi Tang, Xiaolei Wang, Yupeng Hou, Yingqian Min, Beichen Zhang, Junjie Zhang, Zican Dong, et~al. 2023.
\newblock A survey of large language models.
\newblock \emph{arXiv preprint arXiv:2303.18223}.

\bibitem[{Zheng et~al.(2023)Zheng, Chiang, Sheng, Zhuang, Wu, Zhuang, Lin, Li, Li, Xing et~al.}]{zheng2023judging}
Lianmin Zheng, Wei-Lin Chiang, Ying Sheng, Siyuan Zhuang, Zhanghao Wu, Yonghao Zhuang, Zi~Lin, Zhuohan Li, Dacheng Li, Eric Xing, et~al. 2023.
\newblock Judging llm-as-a-judge with mt-bench and chatbot arena.
\newblock \emph{arXiv preprint arXiv:2306.05685}.

\bibitem[{Zhou et~al.(2023)Zhou, Chen, Wan, Wen, Song, Yu, Huang, Peng, Yang, Xiao et~al.}]{zhou2023characterglm}
Jinfeng Zhou, Zhuang Chen, Dazhen Wan, Bosi Wen, Yi~Song, Jifan Yu, Yongkang Huang, Libiao Peng, Jiaming Yang, Xiyao Xiao, et~al. 2023.
\newblock Characterglm: Customizing chinese conversational ai characters with large language models.
\newblock \emph{arXiv preprint arXiv:2311.16832}.

\end{thebibliography}
\bibliographystyle{acl_natbib}

\appendix
\section*{Appendix}
\tcbset{
    breakable,
    left*=10pt, right*=10pt,
    top=0pt, bottom=0pt,
    colback=white!10!white,
    colframe=black!75!black,
    fonttitle=\bfseries\large,
    subtitle style={boxrule=0pt,colback=gray!50!white},
}

\section{Wikidata Queries}

We use the following queries to collect character profiles form Wikidata:

\begin{tcolorbox}[title=Query for collecting human characters]
\small
\begin{lstlisting}[
    language=python,
    breaklines=true,
    basicstyle=\ttfamily
]
PREFIX wdt: <http://www.wikidata.org/prop/direct/>
PREFIX wd: <http://www.wikidata.org/entity/>
PREFIX schema: <http://schema.org/>
PREFIX wikibase: <http://wikiba.se/ontology#>
PREFIX rdfs: <http://www.w3.org/2000/01/rdf-schema#>
SELECT ?person ?label 
(COUNT(DISTINCT(?sitelink)) as ?sites)
WHERE { 
  ?person wdt:P31 wd:Q5 .
  ?sitelink schema:about ?person .
  ?person rdfs:label ?label .
  FILTER (lang(?label) = "zh") .
  ?person schema:description ?description.
  FILTER(LANG(?description) = "zh") .
  }
GROUP BY ?person ?label
ORDER BY DESC(?sites)
LIMIT 5000
\end{lstlisting}
\end{tcolorbox}

\begin{tcolorbox}[title=Query for collecting virtual characters]
\small
\begin{lstlisting}[
    language=python,
    breaklines=true,
    basicstyle=\ttfamily
]
PREFIX wdt: <http://www.wikidata.org/prop/direct/>
PREFIX wd: <http://www.wikidata.org/entity/>
PREFIX schema: <http://schema.org/>
PREFIX wikibase: <http://wikiba.se/ontology#>
PREFIX rdfs: <http://www.w3.org/2000/01/rdf-schema#>
SELECT ?person ?label 
(COUNT(DISTINCT(?sitelink)) as ?sites)
WHERE { 
  ?person wdt:P31 wd:Q15632617 .
  ?sitelink schema:about ?person .
  ?person rdfs:label ?label .
  FILTER (lang(?label) = "zh") .
  ?person schema:description ?description.
  FILTER(LANG(?description) = "zh")
  }
GROUP BY ?person ?label
ORDER BY DESC(?sites)
LIMIT 5000
\end{lstlisting}
\end{tcolorbox}

\section{Prompts}

We use the following prompt in dialogue simulation to prompt chat models generate queries and corresponding responses.

\subsection{Query Simulation}\label{app:query_simulation_prompt}

\begin{tcolorbox}[title=Query Simulation]
\small
\begin{lstlisting}[
    breaklines=true,
    basicstyle=\ttfamily
]
You are skilled at designing questions for specific characters based on background information, as follows you will be provided with information for two characters:

[Character A]
The name is {label1}, the description is {description1}, and the aliases also include {aliases1}.
Here are the properties of Character A:
{claims1}
Here is an introduction to Character A:
{wiki1}

[Character B]
The name is {label2}, the description is {description2}, and the aliases also include {aliases2}.
Here are the properties of Character B:
{claims2}
Here is an introduction to Character B:
{wiki2}

Please design 3 questions that Character A can answer, but are not suitable for Character B to answer. The questions should strictly conform to Character A's era background and character setting, but go beyond the era, genre, occupation, age, knowledge, etc., settings of Character B, therefore Character B cannot answer them. Provide an explanation with each question, explaining why Character A can answer it but Character B cannot.

Please use as casual language as possible to ask questions, and try to use the second person for questioning, such as "Who are you?". Please response in English. Please return the results in the following JSON structure:
[{{"question": str}}]
\end{lstlisting}
\end{tcolorbox}

\subsection{Response Simulation}\label{app:response_simulation_prompt}

\begin{tcolorbox}[title=Response Simulation]
\small
\begin{lstlisting}[
    breaklines=true,
    basicstyle=\ttfamily
]
Please answer the questions according to your identity! When encountering questions that do not match your identity, please refuse to answer the question in the role of {label}, and explain the reason for refusal step by step based on your identity. Please do not step out of your role! Please avoid repeatedly restating your identity or name.
    
You are {label}, your description is {description}, and your aliases also include {aliases}.
Here are your properties:
{claims}
Here is your introduction:
{wiki}
\end{lstlisting}
\end{tcolorbox}

\section{Hyperparameters}\label{app:hyperparameters}
\xhdr{Training} We train all models for 300 steps with the 128 global batch size. We set the training sequence length to 8,192. The learning rate is 2e-6, and the minimum learning rate is 2e-7. We mask prompts and ChatML roles during the training.

\xhdr{Inference} We infer all models with topP 0.8, length penalty 1.1, sequence length 8,192, and max new token 2,048. We generate three rounds for query simulation and randomly select one format-valid query, as smaller LLMs sometimes do not follow the output format. And we only generate one response in the response simulation.

\section{API Configurations}\label{app:api-config}
Through Together API\footnote{\url{https://api.together.xyz/playground}}, we infer our open-sourced general baselines, including OpenChat-3.5, Mistral-7B-Instruct-v0.2, and Mixtral-8x7B-Instruct-v0.1.
We use the default hyper-parameters inherently set in the APIs and set the max length to 8,192.
We run our proprietary general baselines through their APIs.
We call Claude2.1 and Wenxin 4.0 APIs with default parameters and OpenAI APIs for GPT-3.5-Turbo~(gpt-3.5-turbo-1106), GPT-4~(gpt-4), GPT-4-Turbo~(gpt-4-1106-preview) with $0.7$ temperature and $8,192$ max length.

As for the role-play expertise baselines, we call CharacterGLM\footnote{\url{https://maas.aminer.cn/dev/api\#characterglm}} through the official API provided by Zhipu AI.
We infer Xingchen through their official Python SDK with default inference parameters\footnote{\url{https://xingchen.aliyun.com/xingchen/document/python_sdk_static_role}}.
For both models, we set the ``user\_name'' and ``user\_info'' as ``user''.
\label{sec:appendix}

\end{document}